\documentclass[10pt,twocolumn,letterpaper]{article}

\usepackage{cvpr}              

\usepackage[dvipsnames]{xcolor}

\definecolor{cvprblue}{rgb}{0.21,0.49,0.74}
\usepackage[pagebackref,breaklinks,colorlinks,citecolor=cvprblue]{hyperref}

\usepackage{times}
\usepackage{epsfig}
\usepackage{graphicx}
\usepackage{amsmath}
\usepackage{amssymb}
\usepackage{multirow}
\usepackage{booktabs}
\usepackage{gensymb}
\usepackage{color,xcolor}
\usepackage{cite}

\usepackage{xspace}
\usepackage{epstopdf}
\usepackage{algorithm}
\usepackage{algpseudocode}

\usepackage{diagbox}
\usepackage{amsmath,amsfonts}
\usepackage{amsthm}
\usepackage{amssymb,amsopn}
\usepackage{bm} 

\newcommand{\vct}[1]{\boldsymbol{#1}} 
\newcommand{\mat}[1]{\boldsymbol{#1}} 

\newcommand{\T}{^{\textrm T}} 

\newcommand{\eat}[1]{}

\usepackage{booktabs}

\graphicspath{ {./figures/} }

\newlength\savewidth\newcommand\shline{\noalign{\global\savewidth\arrayrulewidth
  \global\arrayrulewidth 1pt}\hline\noalign{\global\arrayrulewidth\savewidth}}

\newcommand{\tabincell}[2]{\begin{tabular}{@{}#1@{}}#2\end{tabular}}

\newcommand{\methodname}{{PIP}\xspace}

\renewcommand{\paragraph}[1]{\vspace{1.25mm}\noindent\textbf{#1}}

\usepackage[capitalize]{cleveref}
\crefname{section}{Sec.}{Secs.}
\Crefname{section}{Section}{Sections}
\Crefname{table}{Table}{Tables}
\crefname{table}{Tab.}{Tabs.}

\renewcommand{\paragraph}[1]{\noindent\textbf{#1}\hspace{2mm}}

\usepackage{amsmath}
\newcommand{\degradprompt}{\mat{d}}
\newcommand{\restoreprompt}{\mat{B}}
\newcommand{\univerprompt}{\mat{U}}

\def\confName{CVPR}
\def\confYear{2024}

\title{
Prompt-In-Prompt Learning for Universal Image Restoration}

\author{Zilong Li$^{1}$\qquad Yiming Lei$^{1}$\qquad Chenglong Ma$^{2}$\qquad Junping Zhang$^{1}$\qquad Hongming Shan$^{2,3*}$ \\
$^{1}$ Shanghai Key Lab of Intelligent Information Processing, School of Computer Science,\\
Fudan University, Shanghai 200433, China\\
$^{2}$ Institute of Science and Technology for Brain-inspired Intelligence and MOE Frontiers Center \\for Brain Science,  Fudan University, Shanghai 200433, China\\
$^{3}$ Shanghai Center for Brain Science and Brain-inspired Technology, Shanghai 200031, China\\
{\tt\small \{zilongli23, clma22\}@m.fudan.edu.cn,\quad \{ymlei, jpzhang, hmshan\}@fudan.edu.cn} 
}

\begin{document}
\maketitle
\begin{abstract}
Image restoration, which aims to retrieve and enhance degraded images, is fundamental across a wide range of applications.
While conventional deep learning approaches have notably improved the image quality across various tasks, they still suffer from (\textbf{i}) the high storage cost needed for various task-specific models and (\textbf{ii}) the lack of interactivity and flexibility, hindering their wider application.
Drawing inspiration from the pronounced success of prompts in both linguistic and visual domains, we propose novel Prompt-In-Prompt learning for universal image restoration, named \methodname. 
First, we present two novel prompts, a degradation-aware prompt to encode high-level degradation knowledge and a basic restoration prompt to provide essential low-level information.
Second, we devise a novel prompt-to-prompt interaction module to fuse these two prompts into a universal restoration prompt.
Third, we introduce a selective prompt-to-feature interaction module to modulate the degradation-related feature. 
By doing so, the resultant \methodname works as a plug-and-play module to enhance existing restoration models for universal image restoration.
Extensive experimental results demonstrate the superior performance of  \methodname on multiple restoration tasks, including image denoising, deraining, dehazing, deblurring, and low-light enhancement.
Remarkably, \methodname is interpretable, flexible, efficient, and easy-to-use, showing promising potential for real-world applications.
The code is available at \url{https://github.com/longzilicart/pip_universal}.

\end{abstract}
    
\section{Introduction}
\label{sec:intro}

\begin{figure}[!htb]
    \centering
    \includegraphics[width=1\linewidth]{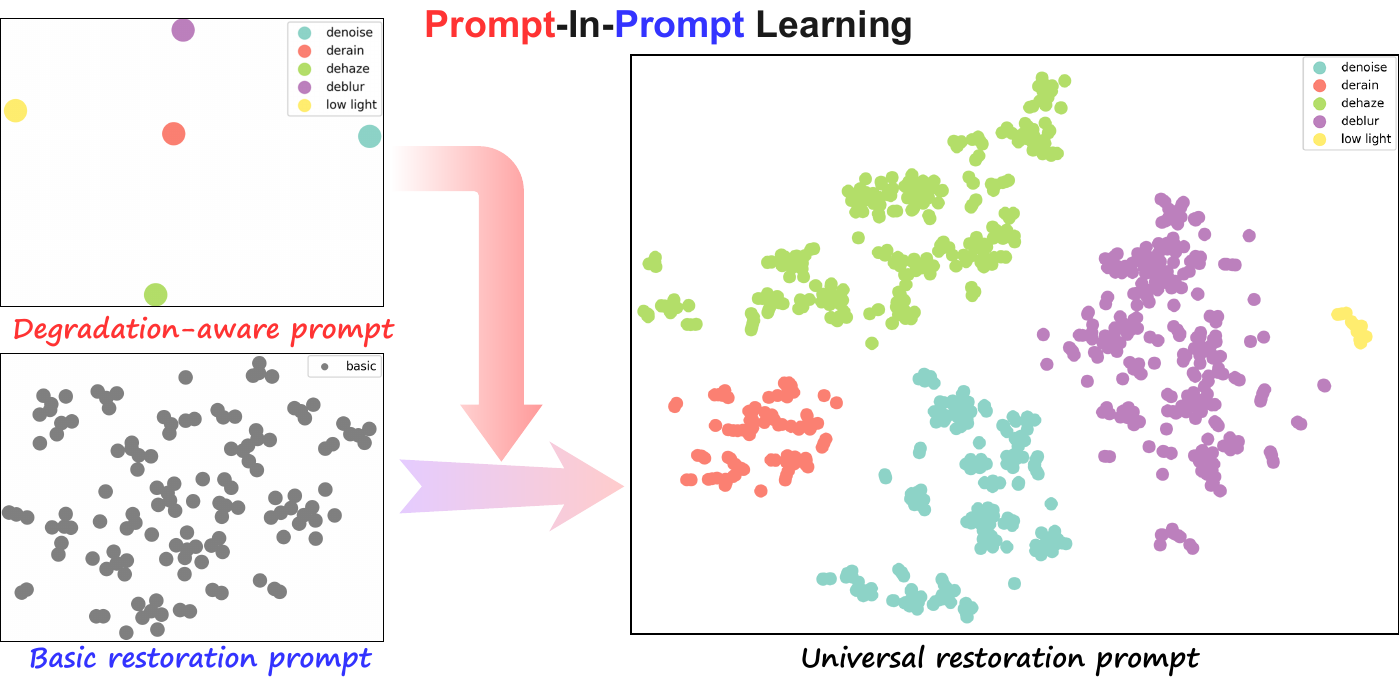}
    \vspace{-15pt}
    \caption{$t$-SNE of the proposed prompt. Prompt-in-prompt learning combines degradation-aware and basic restoration prompts for high- and low-level knowledge simultaneously. The resulting universal restoration prompt is interpretable, offering decoupled properties for different degradation types while being effective for restoration models.}
    \label{fig:motivation} 
    \vspace{-10pt}
\end{figure}

Image restoration aims to faithfully restore the high-quality clean image from various degradations (\eg, noise, rain drops, haze) encountered during image acquisition~\cite{tpami_degradation, image_restoration1, image_restoration2}. Due to the illness nature of the restoration task, a degraded image can be explained by multiple plausible ``clean solutions''~\cite{noisetonoise, deepimageprior}. Consequently, restoration poses significant challenges. 
Various deep learning methods have been developed~\cite{restormer,swinir} and shown effectiveness in specific restoration tasks.
However, in practical scenarios, degradation often manifests in complex forms, and may involve various degradation types (\eg, noise and rain) simultaneously. 
Directly applying task-specific models leads to high storage costs, limited flexibility and generalizability, thereby underscoring the need of universal restoration models.

Recently, research is shifting towards addressing multiple degradations with a single model. A straightforward approach is to modulate the parameter space for each degradation type. This includes designing additional restoration heads for different tasks~\cite{ipt}, integrating degradation-specific paths or modules using contrastive learning~\cite{whether_universal, AirNet}, and fine-tuning degradation-specific parameters identified by explanation methods such as integrated gradients~\cite{FAIG, FAIG_universal}. 
Although these approaches enhance model adaptability to a broader range of scenarios, they still require task-specific training and additional parameters, and may fall short in offering a comprehensive solution.
Prompt learning, on the other hand, enhances the input data with specific ``conditions'', demonstrating impressive flexibility across different domains, and is widely used in large language models and conditional image generation. Pioneering studies like ProRes~\cite{prores} and PromptIR~\cite{promptir} have explored using prompts to adopt models for various degradation tasks, revealing the potential of prompt learning in the field of universal image restoration.

\paragraph{Motivation.}
By reviewing the design philosophy and rationale behind the success of prompts, we find that by manipulating the input data, prompts enable a single model to effectively navigate and utilize its extensive parameter space to meet the needs of different tasks. In essence, prompts affect the input to align with the model's trained knowledge base.
When focusing on restoration tasks, prompts should assist the model in at least three key aspects. \emph{First}, prompts should clearly identify and illustrate the overall state of degradation and navigate the model at a high level. 
\emph{Second}, it should also highlight the low-level details that are relevant to the specific type of degradation and facilitate the restoration model in addressing such degradation. \emph{Third}, prompts should modulate the feature effectively and appropriately to yield optimal results for a specific task.

\paragraph{Prompt-in-prompt learning.} 
The underlying principle of prompt inspires us to design prompt-in-prompt (\methodname) learning. Specifically, \methodname learns two types of prompts to generate the final universal restoration prompts, as illustrated in \cref{fig:motivation}. 
Imitating the decoupled nature of human language, \methodname first learns well-defined degradation-aware prompt to represent specific degradation types and concepts, which serve as a high-level condition to modify the input. 
Nonetheless, degradation-aware prompt may not be sufficient to guide the restoration process directly~\cite{prores}, while the degradation patterns are hard to describe via simple words or conditions and require detailed and textured representations. 
Hence, we propose to learn a basic restoration prompt for key low-level features of various degradations, including essential textures and fine structures, which are more compatible with restoration models. 
Given these two types of prompts, we then design a novel \emph{prompt-to-prompt (P2P) interaction} module to fuse them to generate the final universal restoration prompt, which encodes both rich semantics of degradation and essential detailed information for restoration. 
Unlike linguistic prompts with distinct word meanings, restoration prompts often contain extraneous information, reflecting the complexity of image degradation. For example, rain within an image may exhibit different directional shifts (left or right), posing challenges in formulating precise prompts. 
Hence, instead of directly utilizing the generated prompts to interact with the features, we further introduce  \emph{selective prompt-to-feature (P2F) interaction} to focus only on the most effective features denoted by the attention map.

With the core design mentioned above, \methodname works as a plug-and-play module to enhance existing single-task models for universal restoration. 
While \methodname can be integrated into various positions of a network, we propose to apply it exclusively to the skip connections of the prevalent U-shape networks in the field of image restoration.
This is because the skip connections contribute mainly to those high-frequency details that differ significantly among tasks, where \methodname can better unlock its potential without much computational cost.

\paragraph{Contributions.} Our contributions are as follows.
\begin{itemize}
    \item We propose prompt-in-prompt learning for universal image restoration, which involves learning high-level degradation-aware prompts and low-level basic restoration prompts simultaneously.
    \item We devise a prompt-to-prompt interaction module to fuse these two prompts for a universal restoration prompt. 
    \item We introduce a selective prompt-to-feature interaction module to modulate the most degradation-related features for a specific restoration task. 
    \item Extensive experimental results across various restoration tasks demonstrate the superior performance of our \methodname. We also highlight that \methodname is interpretable, flexible, efficient, and easy to use.
\end{itemize}

\section{Related work}
\label{sec:related_work}

\begin{figure*}[!htb]
    \centering
    \includegraphics[width=0.9\linewidth]{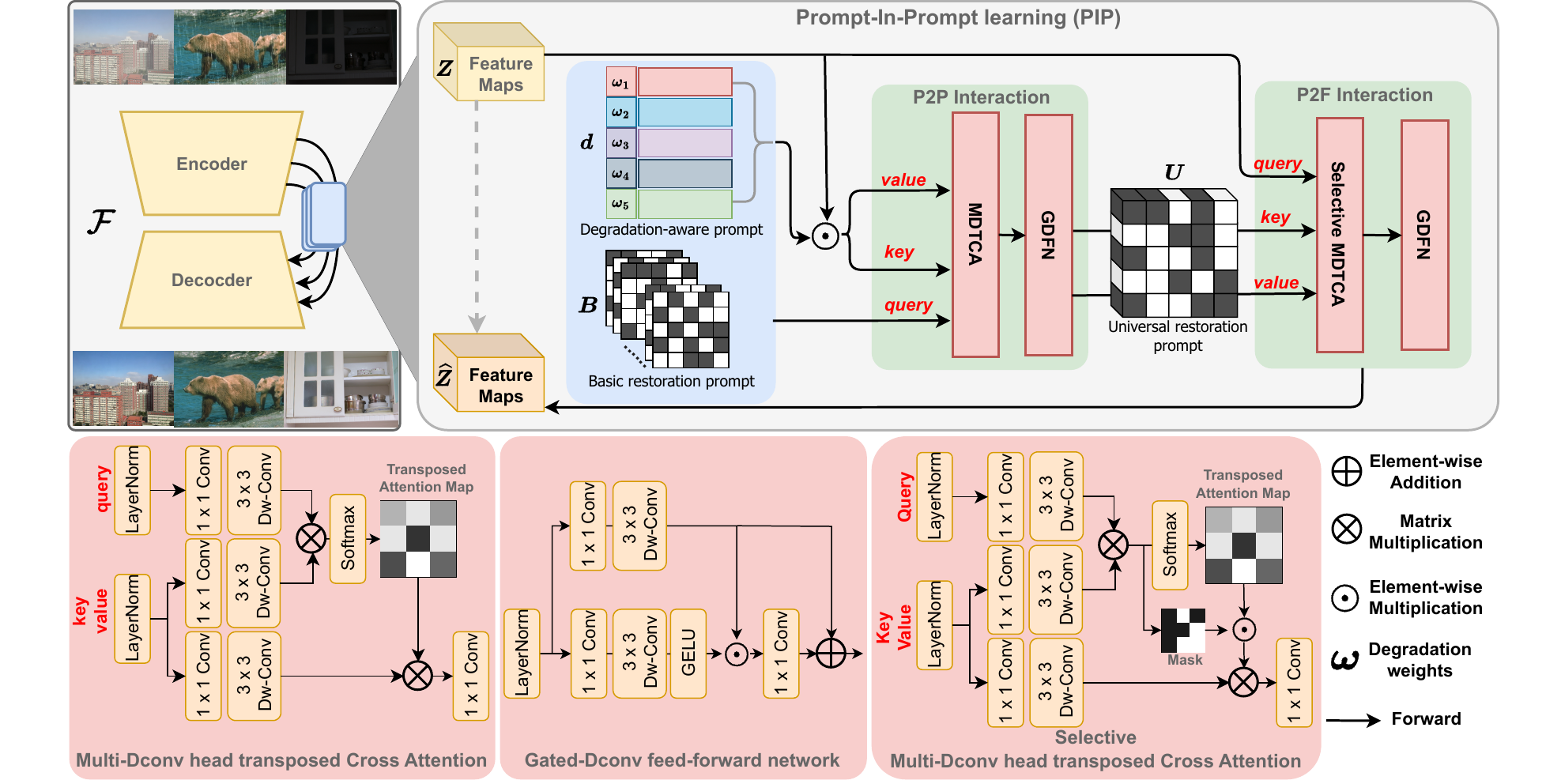}
    \caption{
    The architecture of \methodname, which is developed as a lightweight, plug-and-play module on the skip connection to enhance current task-specific networks for universal restoration. \methodname first learns degradation-aware prompt $\degradprompt$ and basic restoration prompt $\restoreprompt$, and then generates the universal restoration prompt $\univerprompt$ by prompt-to-prompt (P2P) interaction. Lastly, features $\mat{Z}$ are modulated via selective prompt-to-feature (P2F) interaction, focusing on the most relevant features denoted by the prompt. 
    }
    \label{fig:flow} 
\end{figure*}

\paragraph{Multi-task restoration network.}
Image restoration aims to recover clean images or signals from their degraded version, which is significantly different for each task. 
To this end, the literature proposes various methods for an individual degradation task by fully considering the degradation prior~\cite{hinet, mprnet,restormer,swinir, denoise_sota, multi_stage_mlp, deconv,deblur_samplugin,uncertain_deblur,topkderain,uncertain_deblur,deblur_samplugin}. 
Compared with single-task restoration, multi-task restoration is more applicable and advantageous in model storage efficiency.
The primary challenge lies in using a single model to handle various types of degradation and accordingly restore the specific components. 
One solution is to modify the parameter space to fit the model for different degradations~\cite{dasr, AirNet, FAIG_universal}. 
For example, IPT~\cite{ipt} proposes to utilize a pre-trained transformer backbone with different encoders and decoder heads to restore various degradation images.
DASR~\cite{dasr} and AirNet~\cite{AirNet} propose to learn discriminative degradation representation, which is then used to guide the restoration process. 
ADMS~\cite{FAIG_universal} proposes to utilize filter attribution integral gradient (FAIG)~\cite{FAIG} to find the most discriminative parameters related to different components to learn another set of these parameters to better fit different tasks. 
Given the complexity of degradation, these approaches lack flexibility and can still incur considerable storage costs.

Prompt, on the other hand, can be considered as a well-known prior recognized by the model and thus conditioning inputs for ideal outputs. 
Motivated by its easy implementation and high generalization, pioneering studies like ProRes~\cite{prores} and PromptIR~\cite{promptir} have utilized prompts to extend models to various degradations. Specifically, ProRes proposes adding a learnable prompt in the input image, while PromptIR uses various prompts to modulate the feature map. 
Although they have demonstrated potential in enhancing restoration performance with prompts, there remains a substantial disparity between the performance and the controllability exhibited by prompts in other tasks, such as image generation and editing~\cite{cot,smartbrush_inpainting,stablediffusion}.
\emph{This paper develops a lightweight plug-and-play module named \methodname, which enhances existing backbones for state-of-the-art performance in universal restoration while providing flexible interaction and interpretability.} 
Please refer to Sec.~\textcolor{red}{A} in Supplementary Material for detailed related work.

\paragraph{Prompt learning in vision.}
Prompt learning has emerged as a pivotal technique in the field of large language models (LLMs)~\cite{lu2022prompt, promptsurvey}. Research indicates that incorporating prompts into the input can significantly enhance model performance, offering exciting zero-shot or few-shot ability. The primary benefit of using prompts lies in their simplicity and flexibility. By merely ``prompting'' the model with some additional inputs, LLMs can be effectively adapted to a wide range of tasks and scenarios, which is particularly valuable in practical applications~\cite{cot, visual_cot}. 
Vision prompts have also shown promise across multiple tasks, particularly in AIGC tasks such as image generation, inpainting, and editing. Notably, studies have expanded prompts to include vision priors such as segmentation masks, points, or anchor boxes~\cite{segment_anything, smartbrush_inpainting, stablediffusion_inpainting, imagen, restore_SAM_anything}. However, most of them still focus on high-level vision tasks.

By reviewing the philosophy of prompt learning from an image restoration perspective, this paper proposes \emph{prompt-in-prompt learning}, which simultaneously learns high-level degradation-aware knowledge and essential low-level information to prompt existing restoration models. Notably, our design may also benefit other low-level tasks with prompts.

\section{Method}
\label{sec:method}

\subsection{Problem Definition}

Image restoration aims to reconstruct a clean image $\mat{Y}$, from its degraded counterpart $\mat{X}$, represented as $\mat{X} = \mathcal{D}(\mat{Y})$, where $\mathcal{D}$ denotes the degradation process such as noise, rain, haze, \etc. Given a degraded image, conventional approaches employ separate task-specific model $\mathcal{F}_\mathcal{D}$ for each degradation $\mathcal{D}$:
\begin{align}
    \widehat{\mat{Y}} = \mathcal{F}_\mathcal{D}(\mat{X}),
\end{align}
where $\widehat{\mat{Y}}$ is the restored image. In the real restoration scenarios, different degraded images may exhibit, \eg, $\mat{X}_\text{noise}$, $\mat{X}_\text{rain}$, and $\mat{X}_\text{haze}$. Simply implementing task-specific models is unsuitable, resulting in high storage costs and limited flexibility. 
Recognizing the intricacies of practical scenarios, the demand for a universal restoration model $\mathcal{F}$ has surged, which aims at tackling different degradations and consistently generating high-quality restored outputs:
\begin{align}
    \widehat{\mat{Y}} = \mathcal{F}(\mat{X}).
\end{align}

\subsection{Overview of \methodname}

\cref{fig:flow} presents the overview of our \methodname, which works as a plug-and-play module applied to the skip-connections of a U-shape encoder-decoder network. 
The key components in \methodname are two prompts: the degradation-aware prompt to encode high-level degradation knowledge and the basic restoration prompt to provide essential low-level information. These two prompts are then fused to produce the universal restoration prompt through a prompt-to-prompt interaction module, which is used to further modulate the most degradation-related features through a selective prompt-to-feature interaction module.

Concretely, \methodname modulates the network $\mathcal{F}$ on latent features $\mat{Z}$ for different degradations with degradation weights $\vct{\omega}$. The modulated features $\widehat{\mat{Z}}$ is given by:
\begin{align}
    \widehat{\mat{Z}} = \texttt{\methodname}(\mat{Z}, \vct{\omega}),
\end{align}
where $\vct{\omega} = [{\omega}_{1}, {\omega}_{2}, \ldots, {\omega}_{T}]$ denotes the degradation-specific weights and $T$ is the number of degradation types. Notably, we keep $\vct{\omega}$ as the unique controllable interface of \methodname to provide flexible interaction by either human or degradation-assessment models.

In the following, we details these two prompts in \methodname.
\paragraph{Degradation-aware prompt.} 
In the realm of the large language models, the prompt is known for its remarkable flexibility and controllability, which facilitates user interaction and provides meaningful interpretability~\cite{cot, visual_cot}. 
In low-level vision, however, describing the degradation with a few words is difficult due to the complexity of degradation. Finding an effective way to incorporate the degradation concept into models remains challenging. 
One essential design of \methodname is to learn a group of decoupled, degradation-aware prompts  during training. These prompts are learned to clearly represent the concepts of different degradation types, akin to the clarity of human languages. This provides clear interpretation and enables seamless interaction and control by humans and degradation-aware models.

Specifically, we use a 1D vector $\vct{d}_{t}$  to denote the degradation-aware prompt for $t$-th  degradation type, and all prompts, $\{\vct{d}_{t} \}_{t=1}^{T}$, should be directionally decoupled. 
To achieve this, we conduct a directional decoupled loss among them, which is defined as: 
\begin{align}
    \mathcal{L}_\text{ddl} &= \frac{2}{T(T-1)} \sum_{i=1}^{T-1} \sum_{j=i+1}^{T} \max(0, \theta_\text{thre} - \theta_{ij}), \notag \\
     &\text{where}\quad\theta_{ij} = \cos^{-1}\left( \tfrac{\mat{d}_{{i}}\T \mat{d}_{{j}} }{\max(\|\mat{d}_{{i}}\|_2  \|\mat{d}_{{j}}\|_2, \epsilon)} \right),\label{eq:L_pd} 
\end{align}
where the loss $\mathcal{L}_\text{ddl}$ computes the sum of the differences between a threshold angle $\theta_\text{thre}$ and angle $\theta_{ij}$ across all pairs. $\epsilon$ is set as $1 \times 10^{-8}$ to avoid division by zero. This encourages the angles between degradation-aware prompts to be at least $\theta_\text{thre}$ degrees.

Since the degradation types are easy to obtain during training, well-decoupled degradation-aware prompts can be easily learned under supervision. Unlike previous prompt learning that has no explicit learning target~\cite{prores, promptir}, ours  can provide rich knowledge and semantic information of the degradation type through Eq.~\eqref{eq:L_pd}, thereby offering richer interpretability and flexibility.

\paragraph{Basic restoration prompt.}
Although we have degradation-aware prompts denoting the degradation types, it still requires detailed and low-level feature representations for the restoration models. 
Hence, we propose to learn a task-unrelated basic restoration prompt for key low-level features among various degradations, including essential textures and fine structures. 
The basic restoration prompt is denoted as $\restoreprompt \in\mathbb{R}^{c \times h \times w}$, where $c$, $h$, and $w$ denote the channel, height, and width of the prompt, respectively. 
Compared with $\degradprompt$, $\restoreprompt$ may not be easily understood or controlled by humans but is more compatible with restoration models.

Next, we detail how to fuse these two prompts and modulate the features, followed by the training and optimization of \methodname.

\subsection{Prompt-to-Prompt Interaction}
The magic of prompts in vision-language models can be largely attributed to the pre-trained vision-language models such as CLIP~\cite{clip} and BLIP~\cite{BLIP}, which link concepts across different modalities in a shared space.

For image restoration, fusing the degradation-aware prompt $\degradprompt$ and basic restoration prompt $\restoreprompt$ is also important. Therefore, we design a prompt-to-prompt interaction module, using $\degradprompt$ as prompts to guide the generation of the final universal restoration prompt from $\restoreprompt$. 
Specifically, we multiply degradation-aware prompt $\degradprompt$ with the weights $\vct{\omega}$, and then perform element-wise multiplication with the output and $\mat{Z}$ along the channel dimension.  
Then, the output is reshaped and repeated on the width and height to the same shape of the basic restoration prompt $\restoreprompt$ to obtain $\widehat{\mat{D}}$.  
Finally, these two prompts are fused via cross attention mechanism~\cite{cross_atten,prompttoprompt,stablediffusion}, which is defined as:
\begin{align}
    \texttt{CAtt}(\mathbf{Q}_\text{b}, \mathbf{K}_\text{d}, \mathbf{V}_\text{d}) &= \operatorname{softmax}\left(
    {\mathbf{Q}_\text{b} \mathbf{K}_\text{d}\T} / {\sqrt{d_k}}\right) {\mathbf{V}_\text{d}},
\end{align}
where the query $\mathbf{Q}_\text{b}$ is derived from 
the basic restoration prompt $\restoreprompt$, and the key $\mathbf{K}_\text{d}$ and value  $\mathbf{V}_\text{d}$ are derived from $\widehat{\mat{D}}$; in detail, these three matrices are generated through layer normalization, 1$ \times$1 convolutions, and 3$\times$3 depth-wise convolutions as orders, as illustrated in \cref{fig:flow}. 
Since conventional attention mechanism with learnable matrices~\cite{self_attention} can introduce extremely high computation costs and can not well-fit features of different shapes, which are commonly met in image restoration, we adopted the design of Multi-Dconv head Transposed Attention (MDTA)~\cite{restormer} to build our cross attention.  
Then, we employ Gated-Dconv Feed-forward Network (GDFN)~\cite{restormer} to generate the universal restoration prompt $\univerprompt$ for feature modulation based on the attention map and $\restoreprompt$. The prompt-to-prompt interaction can be written as:
\begin{align}
    \univerprompt = \texttt{GDFN}(\restoreprompt + \texttt{CAtt}(\mathbf{Q}_\text{b}, \mathbf{K}_\text{d}, \mathbf{V}_\text{d})).
\end{align}
Finally, the resulting universal restoration prompt $\univerprompt$ serves as the prompt of features with both knowledge of degradation type and can be well understood by restoration models.

\subsection{Prompt-To-Feature Interaction}
As outlined in \cref{sec:related_work}, a key challenge in image restoration lies in addressing various types of degradation with a single model, each focusing on different details. 
For instance, image denoising targets high-frequency details like textures while image deblurring and low-light enhancement prioritize restoring structural and global value shifts. 
To this end, we design prompt-to-feature interaction to modulate the most degradation-related features.

\paragraph{Selective prompt modulation.} 
Although the generated universal restoration prompt $\univerprompt$ can well fit specific degradations, not all the components have an equal contribution to a specific image.
Motivated by sparsity transfromer~\cite{sparse_transformer, topkderain}, we proposed to be more selective during prompt-to-feature interaction to focus on modulating the most degradation-related feature. 
Specifically, we compute the transposed cross attention map $\mat{A}\in\mathbb{R}^{C \times C}$ with channel $C$, where each channel identifies the attention score of $\univerprompt$ and input feature $\mat{Z}$ across the channels, and each row indicates the scores of a typical channel of the feature to the total $C$ prompts. 
Since each channel of feature maps carries a specific meaning of the image, we then select a partial of the attention map in each row using a mask $\boldsymbol{M} \in \mathbb{R}^{C \times C}$ that denotes the most important prompt. 
Similar to previous work~\cite{topkkvt, topkderain}, we select top-$m$ values of each row by applying the mask $\boldsymbol{M}$ defined as follows:
\begin{align}
    \boldsymbol{M}_{i j}= \begin{cases}
    1, & \mat{A}_{i j} \geq \text{Top}_m (\mat{A}_{i\cdot}),\\
    0, & \text{ otherwise.}
    \end{cases}
\end{align}
where $\text{Top}_m (\mat{A}_{i\cdot})$ denotes the $m$ largest values in the $i$-th row of $\mat{A}$. Then, we produce element-wise multiplication on the attention map and mask to compute the cross-attention output, which can be written as: 
\begin{align}
    \texttt{CAtt}_\text{s} (\mathbf{Q}_{\text{z}}, \mathbf{K}_{\text{u}}, \mathbf{V}_{\text{u}}) &= \operatorname{softmax}\left( \boldsymbol{M} \!\odot\! \frac{\mathbf{Q}_{\text{z}} \mathbf{K}_\text{u}\T}{\sqrt{d_k}}\right)  \mathbf{V}_{\text{u}},
\end{align}
where the query $\mathbf{Q}_{\text{z}}$ is derived from feature $\mat{Z}$, and the key  $\mathbf{K}_\text{u}$ and value $\mathbf{V}_{\text{u}}$ are derived from universal restoration prompt $\univerprompt$. $\odot$ represents the element-wise multiplication. 
Lastly, the modulated feature $\widehat{\mat{Z}}$ from the original feature $\mat{Z}$ is given by:
\begin{align}
    \widehat{\mat{Z}} = \texttt{GDFN}(\mat{Z} + \texttt{CAtt}_\text{s}(\mathbf{Q}_{\text{z}}, \mathbf{K}_\text{u}, \mathbf{V}_\text{u})).
\end{align}

\begin{table*}[htb]
    \centering
    \caption{Performance on ``noise-rain-haze'' settings. The best and second best results are marked in \textbf{bold} and \underline{underline}, respectively.}
    \label{tab:compare_three}
    \vspace{-2mm}
    \small
    \begin{tabular}{l | c c c | c | c | c}
    \shline
    \multirow{3}{*}{Method} & \multicolumn{3}{c|}{Denoise (BSD68)} & Dehaze (SOTS) & Derain (Rain100L) & Average \\
    & $\sigma = 15$ & $\sigma = 25$ & $\sigma = 50$ & & & \\
    \hline
    BRDNet~\cite{BRDNet} & 32.26/0.898 & 29.76/0.836 & 26.34/0.836 & 23.23/0.895 & 27.42/0.895 & 27.80/0.843 \\
    LPNet~\cite{LPNet} & 26.47/0.778 & 24.77/0.748 & 21.26/0.552 & 20.84/0.828 & 24.88/0.784 & 23.64/0.738\\
    FDGAN~\cite{FDGAN} & 30.25/0.910 & 28.81/0.868 & 26.43/0.776 & 24.71/0.924 & 29.89/0.933 & 28.02/0.883\\
    MPRNet~\cite{mprnet} & 33.54/0.927 & 30.89/0.880 & 27.56/0.779 & 25.28/0.954 & 33.57/0.954 & 30.17/0.899 \\
    DL~\cite{pami_general_restore} & 33.05/0.914 & 30.41/0.861 & 26.90/0.740 & 26.92/0.391 & 32.62/0.931 & 29.98/0.875 \\
    AirNet~\cite{AirNet} & 33.92/0.933 & 31.26/0.888 & 28.00/0.797 & 27.94/0.962 & 34.90/0.967 & 31.20/0.910 \\
    PromptIR~\cite{promptir} & \underline{33.98}/\underline{0.933} & \underline{31.31}/\underline{0.888} & \underline{28.06}/\underline{0.799} & \underline{30.58}/\underline{0.974} & \underline{36.37}/\underline{0.972} & \underline{32.06}/\underline{0.913} \\
    \hline
    PIP\textsubscript{Restormer} 
    & \textbf{34.24}/\textbf{0.936} & \textbf{31.60}/\textbf{0.893} & \textbf{28.35}/\textbf{0.806} & \textbf{32.09}/\textbf{0.981} & \textbf{38.29}/\textbf{0.984} & \textbf{32.91}/\textbf{0.920} \\
    \shline
    \end{tabular}
\end{table*}

\begin{figure*}[!htb]
    \centering
    \includegraphics[width=1.0\linewidth]{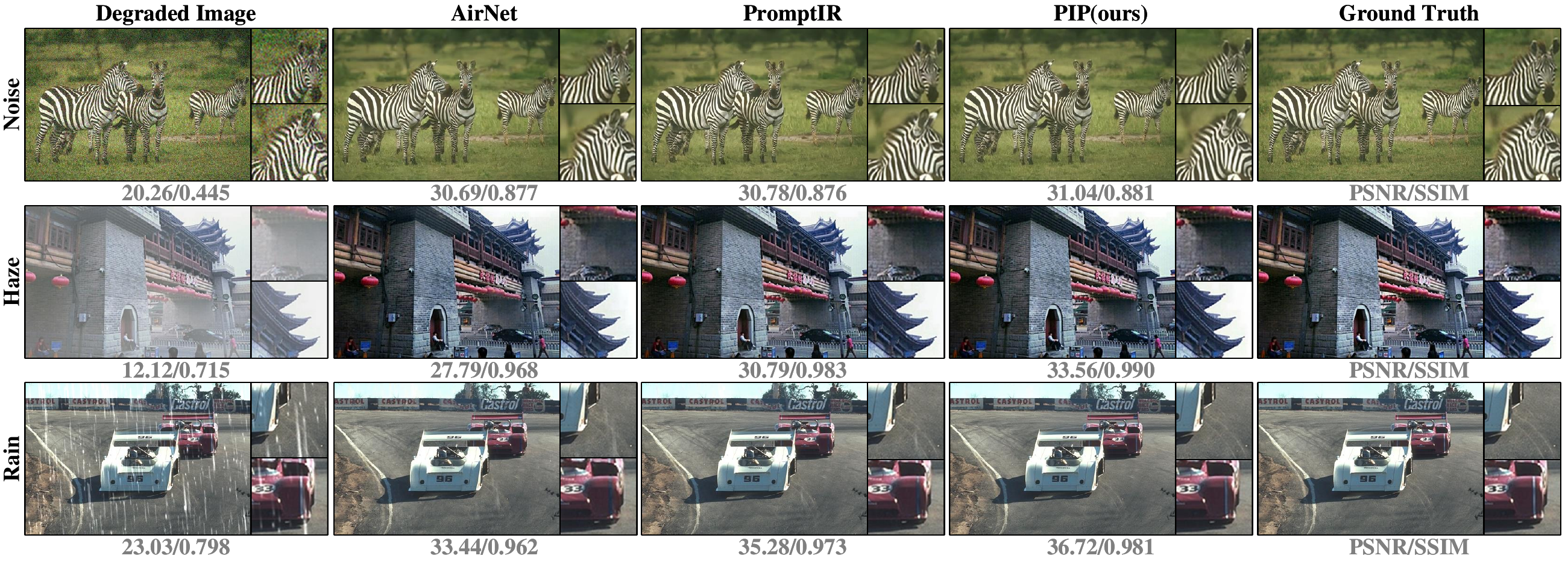}
    \vspace{-20pt}
    \caption{Visual comparison of universal methods on image denoising, deraining, and dehazing.}
    \label{fig:compare_three}
    \vspace{-5pt}
\end{figure*}

\subsection{Training and Optimization}
We train \methodname  on multiple datasets for universal restoration.
Specifically, we integrate \methodname into the skip connection of U-shape backbones, as early-stage features exhibit more significant differences between tasks. 
During the training phase, the degradation weights $\vct{\omega}$ are given as one-hot labels to select the degradation-aware prompt.  We incorporate \methodname on the widely used backbones, including Restormer~\cite{restormer} and NAFNet~\cite{nafnet}. 
To focus on the effect of \methodname, we only involve the most common optimization process, with random horizontal and vertical flips as the data augmentation and pixel-wise $\mathcal{L}_1$ for restoration loss. 
The total loss of \methodname can be written as: 
\begin{align}
    \mathcal{L} = \mathcal{L}_1 + \alpha \mathcal{L}_{\text{ddl}}, 
\end{align}
where $\alpha$ is a trade-off hyperparameter.


\section{Result}

We conduct experiments on five restoration tasks, including image denoising, deraining, dehazing, deblurring, and low-light enhancement.

\subsection{Experimental Setup}
\paragraph{Dataset.}
Regarding image denoising, deraining, and dehazing, we use datasets in line with previous works~\cite{AirNet, promptir}. The training datasets include BSD400~\cite{BSD400} and WED~\cite{ma2016waterloo_wed} for denoising by adding Gaussian noise levels $\sigma \in \{15,25,50\}$,  Rain100L~\cite{yang2020learning_rain100} for deraining, and SOTS~\cite{li2018benchmarking_SOTS} for dehazing. In addition, we include two challenging tasks of image deblurring and low-light enhancement and use the GoPro~\cite{gopro} and LOL dataset~\cite{lol} for training, as previous research~\cite{restormer,nafnet,acmmm_universal}.
For all the datasets, we follow the standard practices in data splitting and preprocessing in the field. 
Finally, we train and evaluate \methodname jointly on multiple datasets, using the three-task configuration ``noise-rain-haze'' and the five-task setup ``noise-rain-haze-blur-enhance''. 
Please refer to Sec.~\textcolor{red}{B} in Supplementary Material for more detailed information about the training and testing datasets.

\paragraph{Implementation details.}
During the training phase, the degradation weights $\vct{\omega}$ are given as one-hot labels to select the degradation-aware prompt in a supervised fashion. 
We incorporate \methodname on the widely used backbones, including Restormer and RAPNet on the skip connections. 
For simplicity, we utilize Adam optimizer with $\beta_1 = 0.9$ and $\beta_2 = 0.99$ to optimize the network. The learning rate is set to $5\times{10}^{-4}$ with a total batch size of 32 for 200 epochs on 8 RTX4090 GPUs. During training, we utilized cropped patches with a size of $128 \times 128$ for 150 epochs and $200 \times 200$ for the rest 50 epochs, using random horizontal and vertical flips as the only augmentation. 
The $\theta_{\text{thre}}$ of $\mathcal{L}_{\text{ddl}}$ is set to 90$^\circ$, and $\alpha$ is set to 0.002. 
In addition to the default settings, we have also developed a data augmentation strategy to introduce multiple degradation types for prompt learning, which is similar to CutMix~\cite{cutmix, cutonce}. This is not involved by default for fair comparison. Please refer to Sec.~\textcolor{red}{B} in  Supplementary Material for more implementation details, such as the training procedure and network configuration.

\paragraph{Evaluation metrics and comparisons methods.} 
We use the peak signal-to-noise ratio (PSNR) and the structural similarity (SSIM)~\cite{ssim} to evaluate the performance. 
We plug \methodname on the skip connection of Restormer~\cite{restormer} and NAFNet~\cite{nafnet} for PIP\textsubscript{Restormer} and PIP\textsubscript{NAFNet}.
Regarding universal restoration methods, we mainly take the state-of-the-art AirNet~\cite{AirNet} and PromptIR~\cite{promptir} for comparison by using the official checkpoint.

\begin{table*}[htb]
    \centering
    \caption{Performance on ``noise-rain-haze-blur-enhance'' datasets. The best and second best results of multi-task restoration are marked in \textbf{bold} and \underline{underline}, respectively. Metrics are presented in [PSNR (dB) / SSIM]. Metrics are presented in [PSNR (dB) / SSIM].}
    \label{tab:compare_five}
    \vspace{-2mm}
    \footnotesize
    \begin{tabular}{ l | l | c c c | c | c | c | c | c}
    \shline
    \multirow{2}{*}{Cat.} & \multirow{2}{*}{Method} & \multicolumn{3}{c|}{Denoise (BSD68)} & Dehaze & Derain & Deblur & Lowlight & \multirow{2}{*}{Average} \\
    & & $\sigma = 15$ & $\sigma = 25$ & $\sigma = 50$ & (SOTS) & (Rain100L) & (GoPro) & (LOL) & \\

    \hline
    \multirow{3}{0.8cm}{Single}
    & SwinIR~\cite{swinir} & 34.42/- & 31.78/- & 28.56/- & - & - & - & - & - \\
    & Restormer~\cite{restormer} & 34.40/- & 31.79/- & 28.60/- & - & 37.61/0.975 & 32.92/0.961 & - & - \\
    & NAFNet~\cite{nafnet} & - & - & - & - & - & 33.69/0.967 & - & - \\
    \hline
    \multirow{3}{0.8cm}{Multi.}
    & Restormer~\cite{restormer} & 33.55/0.926 & 30.96/0.879 & 27.75/0.784 & 29.50/0.973 & 36.53/0.974 & 27.31/0.835 & 23.00/0.841 & 29.94/0.887 \\
    & NAFNet~\cite{nafnet} & 33.71/0.930 & 31.12/0.884 & 27.91/0.793 & 30.67/0.975 & 37.17/0.976 & 28.32/0.863 & 23.18/0.856 & 30.29/0.896  \\
    & PromptIR~\cite{promptir} & 33.67/0.929 & 31.08/0.883 & \underline{27.87}/0.787 & 30.16/0.977 & 37.16/0.977 & \textbf{28.73}/\textbf{0.868} & \underline{23.91}/\underline{0.848} & 30.36/0.895  \\
    \hline
    \multirow{2}{0.8cm}{Ours}

    &    PIP\textsubscript{NAFNet} 
    & \textbf{34.10}/\textbf{0.935}  & \underline{31.45}/\textbf{0.893} & \textbf{28.21}/\textbf{0.806} & \underline{31.75}/\underline{0.978} & \underline{37.67}/\underline{0.980}  & 28.08/0.853 & 23.37/0.854 & \underline{30.66}/\underline{0.899} \\
    &    PIP\textsubscript{Restormer} 
    & \underline{34.05}/\underline{0.934} & \textbf{31.48}/\underline{0.891} & 27.29/\underline{0.805} & \textbf{32.11}/\textbf{0.979} & 
    \textbf{38.09}/\textbf{0.983} & \underline{28.61}/\underline{0.861} & \textbf{24.06}/\textbf{0.859} & \textbf{30.81}/\textbf{0.901} \\
    \shline
    \end{tabular}
\end{table*}

\begin{figure*}[!htb]
    \centering
    \includegraphics[width=1\linewidth]{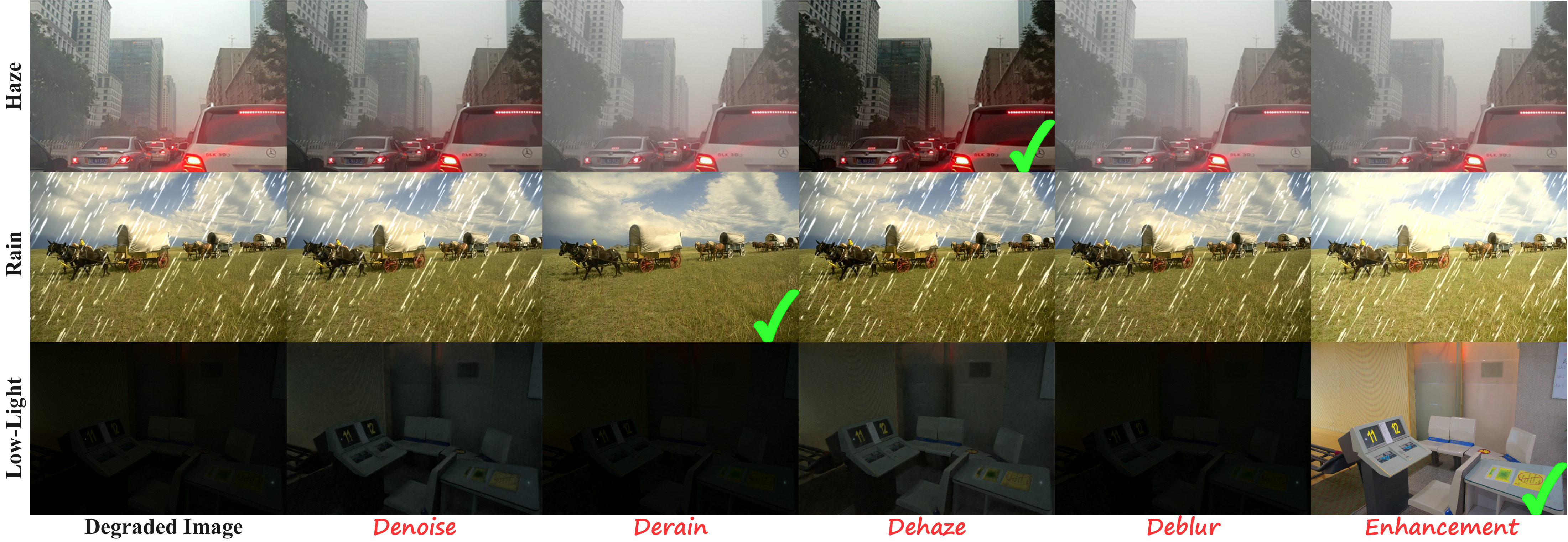}
    \vspace{-15pt}
    \caption{Visualization results of using different degradation-aware prompts for restoration.}
    \label{fig:decouple} 
\end{figure*}

\subsection{Performance on Image Restoration Tasks}

\paragraph{Multi-task performance evaluation.}  
\cref{tab:compare_three} presents the overall performance of \methodname and other state-of-the-art methods on ``noise-rain-haze'' setting.  
We find that our \methodname outperforms the state-of-the-art PromptIR~\cite{promptir} with the same backbone of Restormer~\cite{restormer} across various tasks, demonstrating the effectiveness of our \methodname. 
Notably, \methodname demonstrates significant improvements in challenging, severe degradation tasks such as image deraining and dehazing, outperforming PromptIR by over 1.5 dB in PSNR. 
\cref{fig:compare_three} shows the restoration results of various methods for image denoising, deraining, and dehazing. we find that \methodname achieves better visual quality with less noise and artifacts than the other methods. 

In addition, we find that previous methods tend to eliminate those image details similar to the degradation. For instance, in the third row of \cref{fig:compare_three}, both AirNet and PromptIR remove the white tire line that resembles a raindrop, while \methodname successfully retains the details. 
This is largely due to the high-quality universal restoration prompt proposed in this paper, which effectively guides the network to accurately remove those degradation-related patterns.

\paragraph{Enhancing restoration backbone with \methodname.}
We also assess the performance of \methodname over a broader spectrum of degradation types compared to baseline methods, by training and testing \methodname and other comparison methods on the five-task setting.  
This is more challenging due to severe corruption in blurred and low-light images. 
\cref{tab:compare_five} shows the performance of each method. We note that we adopt the results reported in the original papers for the single-task methods. Since most of them are optimized with techniques such as progressive training~\cite{restormer} and may be trained on different datasets, they can be regarded as the upper bound for universal models. 
Through comparison across multi-task methods, we find that \methodname achieves superior results and effectively improves the backbone models for universal restoration. By integrating \methodname into Restormer, there are noticeable performance gains by at least 0.5 dB and up to 1.5 dB in challenging tasks like deblurring and deraining, respectively. Similar observation can be found by comparing PIP\textsubscript{NAFNet} and NAFNet, demonstrating the effectiveness of \methodname on various backbones.  
As expected, \methodname suffers from a performance drop in image denoising and deraining compared to results in \cref{tab:compare_three}, but image dehazing gets a slight improvement.
This may be attributed to the limited parameter numbers for both the backbone networks and \methodname. 
Generally, we find that \methodname is able to effectively handle a broader range of degradations and maintains good performance. In contrast, conventional prompt-based restoration methods such as PromptIR tend to suffer considerable performance drop when dealing with increasing types of degradation compared to \cref{tab:compare_three}.

\begin{table*}[htb]
    \centering
    \caption{Performance on unseen noise level of ($\sigma = 10, 30, 60$) and severe rain conditions from the Rain100H dataset. Additional noise ($\sigma=25$) was added to the rain and haze test sets to generate images with multiple degradations.}
    \label{tab:performance_degradations_real}
    \vspace{-2mm}
    \small
    \begin{tabular}{l | c c c | c | c | c | c}
    \shline
    Method & \multicolumn{3}{c|}{Denoise (Urban100)} & Rain100H & Rain100L + Noise & Haze + Noise & Average \\
     & $\sigma = 10$ & $\sigma = 30$ & $\sigma = 60$ & & &  \\
    \hline
    AirNet~\cite{AirNet}     
    & 35.26/0.952 & 30.04/0.880  & 25.41/0.714  & 14.55/0.478 & 24.93/0.768 & 15.94/0.753 & 24.35/0.757 \\
    PromptIR~\cite{promptir}  
    & \underline{35.77}/\underline{0.958} & \underline{30.66}/\underline{0.901} & \underline{27.03}/\textbf{0.817}  & \underline{15.54}/\underline{0.486} & \underline{24.92}/\underline{0.767} & \underline{15.93}/\underline{0.752} & \underline{24.97}/\underline{0.780} \\
    PIP\textsubscript{Restormer}
    & \textbf{36.17}/\textbf{0.960} & \textbf{31.24}/\textbf{0.909} & \textbf{27.30}/\underline{0.809}  & \textbf{16.29}/\textbf{0.501} & \textbf{26.50}/\textbf{0.771} & \textbf{16.58}/\textbf{0.754} & \textbf{25.68}/\textbf{0.784} \\
    \shline
    \end{tabular}
\end{table*}

\begin{table*}[htb]
    \centering
    \small
    \caption{Efficiency comparison, including number of parameters, FLOPS, and average inference time. The testing is conducted on a single RTX2080Ti GPU using $1000$ images with a batch size of 1, each at a resolution of $256 \times 256$. }
    \begin{tabular}{l|ccc}
    \shline
    Method & Param. (M) & FLOPS (G) & Infer. (ms) \\
    \hline
    AirNet~\cite{AirNet}        & 5.767 & 301.27 & 417.43  \\
    \hline
    Restormer~\cite{restormer} & 26.09 & 140.99 & 225.87  \\
    PromptIR~\cite{promptir} & 32.96 ($\uparrow$26.32\%) & 158.14 ($\uparrow$12.16\%) & 251.98  ($\uparrow$11.56\%) \\
    PIP\textsubscript{Restormer} & 26.82 ($\uparrow$\textbf{2.77}\%) & 154.72 ($\uparrow$\textbf{9.74}\%) & 245.33  ($\uparrow$\textbf{8.62}\%) \\
    \shline
    \end{tabular}
    \label{tab:efficiency}
\end{table*}

\begin{figure}[!htb]
    \centering
    \includegraphics[width=1\linewidth]{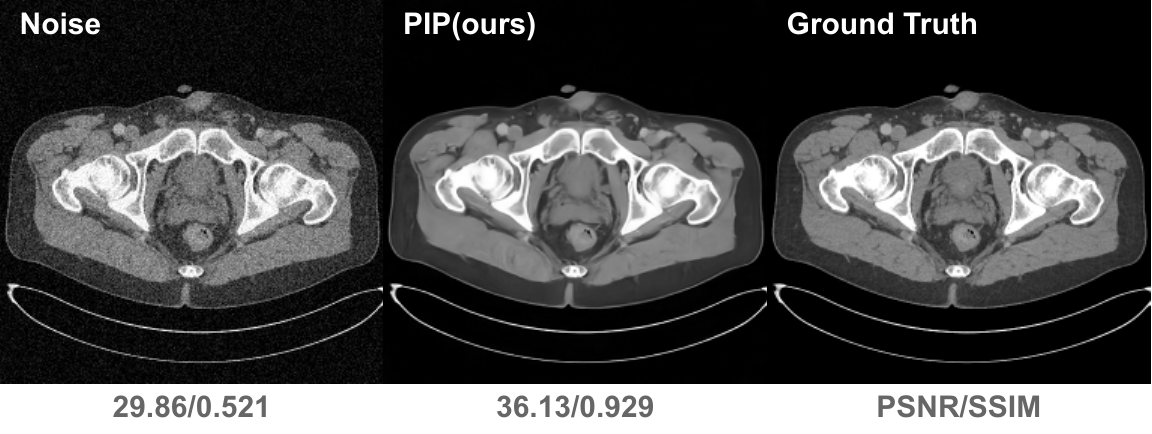}
    \vspace{-20pt}
    \caption{Visualization of \methodname on CT denoising with noise level $\sigma = 25$ under a CT window of [-200, 300] HU.}
    \label{fig:med} 
\end{figure}

\paragraph{Performance on out-of-distribution data.}
We further evaluate \methodname and other universal models on out-of-distribution data, as presented in \cref{tab:performance_degradations_real}. Generally, we find that \methodname outperforms other universal methods on various out-of-distribution degradations, particularly when adapting to unseen noise levels of $\sigma \in \{10, 30, 60\}$ and heavy rain degradation on Rain100H~\cite{yang2020learning_rain100}. \methodname also better handles the multiple degradations than comparison methods. 
Additionally, we directly use \methodname for CT image denoising and find \methodname robust, as present in \cref{fig:med}. 
Please refer to Sec.~\textcolor{red}{C} in Supplementary Materials for more quantitative and qualitative results, including zero-shot performance on real-world datasets and medical images. 
we also discuss how to better fit \methodname for multiple degradations by controlling the degradation-aware prompt and training with data augmentation.

\subsection{Efficiency}
\cref{tab:efficiency} showcases the efficiency of various methods in terms of the number of parameters, FLOPS, and average inference time. These results are obtained on a single RTX2080Ti GPU, using a batch size of 1, and were averaged over 1000 images at a resolution of $256 \times 256$. We observe that the proposed \methodname is efficient, causing only a slight increase in parameters and FLOPS but resulting in notable performance gain when compared to the backbone. It also outperforms PromptIR, which significantly increases the number of parameters.
In conclusion, \methodname is lightweight, and only increases a barely amount of extra computation costs while achieving good performance.

\subsection{Ablation Study}

We evaluate the effectiveness of each component under ``noise-rain-haze'' settings, and present the average results in \cref{tab:ablation}. For configurations \textbf{a}), \textbf{b}) and \textbf{d}) without prompt-in-prompt learning, we repeat and resize them to the same shape as the universal restoration prompt. For other configurations without selective prompt-to-feature interaction, we use the Multi-Dconv head transposed Cross Attention to replace the selective one. 

\begin{table}
    \centering
    \small
    \caption{Quantitative evaluation of different configurations of \methodname under ``noise-rain-haze'' datasets. $\degradprompt$ and $\restoreprompt$ are the degradation-aware prompt and basic restoration prompt, respectively.}
    \begin{tabular}{l|ccc|c}
    \shline
    Config. & $\degradprompt$ & $\restoreprompt$ & selective P2F  & Avg. PSNR/SSIM \\
    \hline
    \textbf{a}) & \checkmark &            &            
    &31.85/0.908 \\
    \textbf{b}) &            & \checkmark &                     
    &32.07/0.911 \\
    \textbf{c}) & \checkmark & \checkmark &                        
    &32.62/0.918 \\
    \textbf{d}) &            & \checkmark & \checkmark                       
    &32.15/0.912 \\
    \textbf{e}) & \checkmark & \checkmark & \checkmark  
    &\textbf{32.91}/\textbf{0.920} \\
    \shline
    \end{tabular}
    \label{tab:ablation}
\end{table}

\begin{figure}[t]
    \centering    
    \includegraphics[width=1\linewidth]{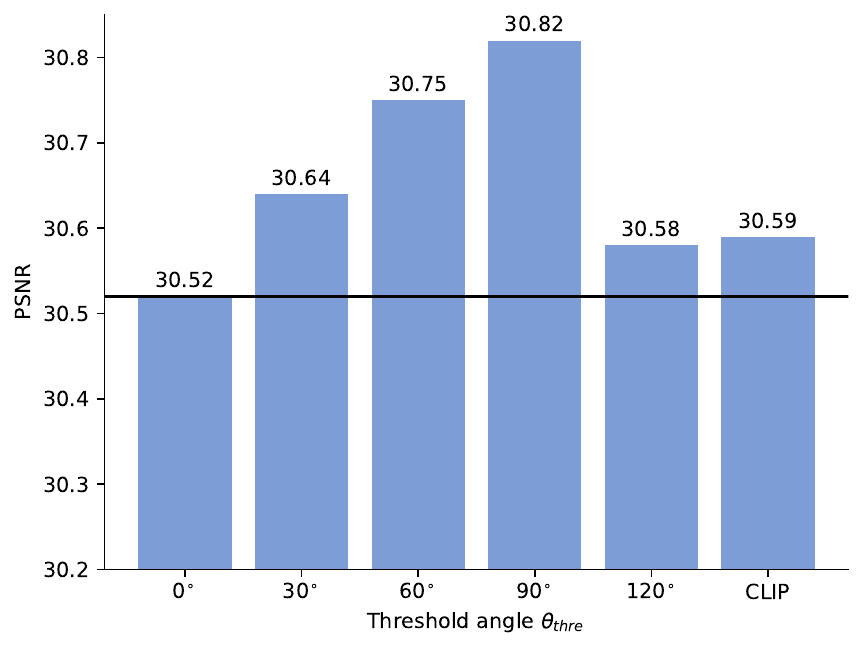}
    \caption{The effect of varying $\theta_\text{thre}$ in $\mathcal{L}_\text{ddl}$ from 0$^\circ$ to 120$^\circ$ during training. The black horizontal line indicates $\mathcal{L}$ with $\alpha = 0$, which imposes no constraint on the degradation-aware prompt; note that it also corresponds to $\theta_\text{thre}=0^\circ$. In the ``CLIP'' configuration, a pre-trained CLIP text encoder~\cite{clip} and a linear layer are used to create degradation-aware prompts of the same shape. The experiments are conducted on a four-dataset setting of ``noise-rain-blur-enhance'' to reduce the computation costs, considering the dehazing dataset is the largest one. 
    }
    \label{fig:suppl_ablationloss} 
\end{figure}

\paragraph{Ablation on prompt-in-prompt learning.} 
Degradation-aware prompts play a crucial role in \methodname by integrating the high-level degradation information into the restoration process. By comparing  configurations \textbf{b}) and \textbf{c}) in \cref{tab:ablation},  we find that the degradation-aware prompts effectively enhance the overall performance. To further assess the effectiveness of the proposed method, we examine \methodname using both accurate and inaccurate prompts. As illustrated in \cref{fig:decouple}, degradation-aware prompts are highly effective and manifest a decoupled behavior, distinct for each degradation type. For instance, only images with the correct prompt, such as ``derain'' in the second row and ``enhance'' in the third row, are successfully restored, and using incorrect prompt results in notable performance drop or restoration failure. 
We also find that relying solely on the degradation prompts leads to sub-optimal results in \textbf{a}), \textbf{b}) and \textbf{c)}. This can be attributed to the misalignment of high-level and low-level knowledge. 
These results highlight the effectiveness of the proposed prompt-in-prompt learning, and the resultant universal restoration prompt can benefit universal restoration in both high- and low-level aspects.

\paragraph{Ablation on selective prompt-to-feature interaction.} 
By comparing \textbf{b}) and \textbf{d}) in Tab.~\ref{tab:ablation}, we find that selective prompt-to-feature interaction is beneficial for restoration. 
This could stem from the imprecise and abundant of restoration prompts while selective interaction can focus on the most pertinent features for a specific degradation. 
Upon comparing \textbf{e}) and \textbf{c)}, both with prompt-in-prompt learning, selective interaction leads to more significant advancements. This is likely due to the improved quality of the universal restoration prompts, which positively guides the selection. 

\paragraph{Ablation on $\mathcal{L}_\text{ddl}$ and $\mathcal{L}$.} 
We explore how different settings of the trade-off hyperparameter $\alpha$ in $\mathcal{L}$ and the threshold angle $\theta_\text{thre}$ in $\mathcal{L}_\text{ddl}$ would affect the restoration performance.
Specifically, $\alpha = 0$ implies no constraints on the learnable degradation-aware prompts, whereas $\theta_{\text{thre}} = 90^\circ$ indicates that the prompts are entirely distinct. As shown in~\cref{fig:suppl_ablationloss}, settings with $\alpha = 0$ and $\theta_\text{thre} = 0^\circ$ are less effective compared to others. This might be due to the entanglement of different degradation-aware prompts, hindering the model from using high-level knowledge. Notably, increasing $\theta_\text{thre}$ improve the performance, with a notable gain of 0.3db when comparing $\theta_\text{thre} = 90^\circ$ to $\theta_\text{thre} = 0^\circ$. This suggests that decoupled degradation-aware prompts can enhance restoration. We also find that $\theta_\text{thre} = 120^\circ$ leads to a performance drop compared to $\theta_\text{thre} = 90^\circ$. In summary, we choose $\theta_\text{thre} = 90^\circ$ for better performance. 

Additionally, we study the efficacy of using a pre-trained CLIP text-encoder~\cite{clip} to generate the high-level degradation-aware prompt. Note that the CLIP text encoder also produces a 1D vector of size 512. We first use degradation classes like ``noise'' or ``blur'' to get language embeddings from CLIP and then project these embeddings through a linear layer to generate degradation-aware prompts of the same shape. 
We find that this approach performs better than configurations without constraints, yet it is still less optimal compared to our default setting of $\theta_\text{thre} = 90^\circ$, as presented in Fig.~\ref{fig:suppl_ablationloss}.

\begin{figure}[!htb]
    \centering
    \includegraphics[width=1\linewidth]{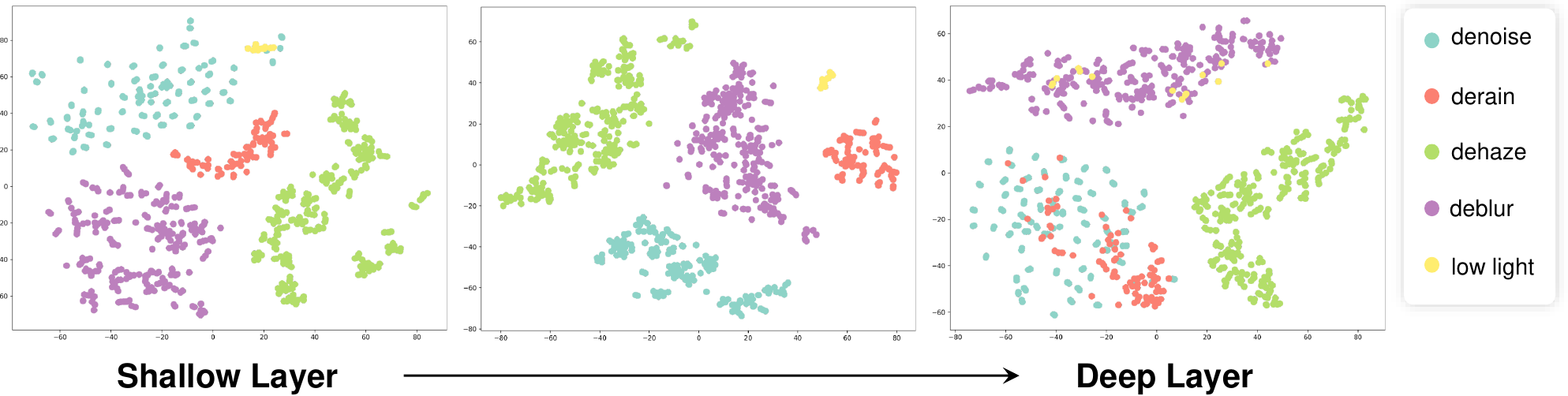}
    \vspace{-15pt}
    \caption{$t$-SNE visualization of the three universal restoration prompt in PIP\textsubscript{Restormer}. Each color represents one degradation type.}
    \label{fig:tsne} 
    \vspace{-10pt}
\end{figure}

\section{Discussion}

\paragraph{Interpretability.} 
We visualize the $t$-SNE of the three generated universal restoration prompts for each skip-connection in \cref{fig:tsne}. 
Interestingly, we find they are well-clustered in terms of the degradation types. In shallow layers, the universal restoration prompt is less compact to fit diverse image details compared to the middle layer. In the deep layer, on the other hand, some degradation types show correlation and cluster together. 
This phenomenon might occur because degradations such as rain and noise, both influenced by high-frequency noise, share similar features in deep layers.
Generally, by learning clear degradation concepts, \methodname is interpretable and flexible. Since \methodname is designed as a plug-and-play module, it is also easy to use and can effectively enhance backbone models.

\paragraph{Control by degradation-aware models.}
\methodname is designed to focus on enhancing the restoration performance rather than automatically recognizing different tasks. This is due to the complicated nature of degradation and various requirements in application. For instance, a user may want to remove noise from a low-light image without increasing brightness. However, we note that \methodname can be \emph{easily controlled via a simple degradation-aware model without performance drop}, referring to the result in Sec.~\textcolor{red}{C} of Supplementary Materials. In general, \methodname is flexible and can easily adapt to different requirements.


\section{Conclusion}

This paper proposes a novel prompt-in-prompt learning for universal image restoration. \methodname involves learning high-level degradation-aware prompts and low-level basic restoration prompts simultaneously to produce the effective universal restoration prompt. By modulating the most degradation-related features with a selective prompt-to-feature interaction module, \methodname achieves superior and robust performance on various restoration tasks. \methodname is also efficient and lightweight, making it easily adaptable to diverse models and enhancing them for universal image restoration.

\clearpage

{
    \small
    \bibliographystyle{ieeenat_fullname}

}

\clearpage
\setcounter{page}{1}
\maketitlesupplementary

\setcounter{figure}{0}
\setcounter{table}{0}
\renewcommand{\thetable}{S\arabic{table}}
\renewcommand{\thefigure}{S\arabic{figure}}
\appendix\label{sec:appendix}

\begin{abstract}
    This supplementary material includes five parts: 
    (\ref{sec:relatedwork}) detailed related work,
    (\ref{sec:implementation}) implementation details,
    (\ref{sec:results}) additional results, 
    and (\ref{sec:discuss}) discussion.
\end{abstract}


\section{Detailed Related Work}
\label{sec:relatedwork}

\paragraph{Image restoration.}
Image restoration aims to recover clean images or signals from their degraded counterparts, which is highly related to the corresponding degradation process. 
For instance, Gaussian noise, typically seen as noise added to an image, mainly affects high-frequency details. On the other hand, degradations such as rain, snow, and haze are associated with atmospheric light or transmission maps~\cite{related_rainsownhaze1, related_rainsownhaze2, related_rainsownhaze3, related_rainsownhaze4, related_rainsownhaze5, related_rainsownhaze6}, which are multiplied to the clean signal. The blurring or low light degradation can be attributed to blur kernel and global value shift, resulting in distorted structure and color. 
Because of the intrinsic distinctions among different degradation processes, various models are proposed to tackle individual degradation tasks~\cite{hinet, mprnet,restormer,swinir, denoise_sota, multi_stage_mlp, deconv,deblur_samplugin,uncertain_deblur}. 
Although incorporating prior information of specific degradation into the model design can yield notable performance improvements, this may compromise generalizability across different tasks or even datasets.

Besides task-specific design, most restoration models share similar fundamental designs, such as using transformers to model long-range dependencies~\cite{swinir,artformer,restormer}, employing multi-scale or pyramid architectures with skip connection to leverage features from various layers~\cite{mprnet, mimounet}, and spatial attention or gating mechanisms that concentrate on degradation-aware features. 
Among them, U-Net~\cite{unet} architectures are particularly popular, which process features at different scales and combine them with skip connections. This helps the network to restore both the structure in deeper layers and the fine details implied by the shadow layers and still demonstrate its superiority in the recent state-of-the-art methods~\cite{restormer, nafnet, deconv, topkderain,uncertain_deblur,deblur_samplugin}.

\section{Implementation Details}
\label{sec:implementation}

\subsection{Dataset and Preparation}
\paragraph{Training and testing datasets.}
We mainly follow the previous works in data preparation. All the datasets used for training and testing are summarized in Tab.~\ref{tab:dataset_summary}. 
For image denoising, we use a combined set of BSD400~\cite{BSD400} and WED~\cite{ma2016waterloo_wed} for training and the noisy image are generated by adding Gaussian noise with different noise levels $\sigma \in \{15,25,50\}$. Testing is performed on BSD68~\cite{martin2001database_bsd} and Urban100~\cite{huang2015single_urban100}. 
For single image deraining, we use the Rain100L~\cite{yang2020learning_rain100} dataset only, which consists of 200 clean-rainy image pairs for training, and 100 pairs for testing. 
Regarding image dehazing, we utilize SOTS dataset~\cite{li2018benchmarking_SOTS} that contains 72,135 training images and 500 testing images. 
For image deblurring, we only train the model on GoPro dataset~\cite{gopro} as previous literature~\cite{restormer,nafnet} with 2103 image pairs for training and 1,111 pairs for testing. 
For low-light enhancement, we utilize LOL~\cite{lol} dataset, which contains 485 and 15 pairs for training and testing, respectively. 

\paragraph{Data preprocessing.}
We follow the data preparation procedure in previous works~\cite{AirNet,promptir}. 
It is clear that the training data is imbalanced across different tasks, particularly with the dataset of dehazing having significantly more data than others. To tackle this, we adopted the same approach as PromptIR~\cite{promptir}, which involves resampling the datasets for other tasks multiple times to match the magnitude of the dehazing data. The resampling ratios are also presented in Tab.~\ref{tab:dataset_summary}. 
All the data are randomly cropped into a size of $128 \times 128$ or $200 \times 200$, with random horizontal and vertical flips as data augmentation.

We acknowledge that additional training data, data augmentation, or larger patch sizes can all be beneficial for the training, especially for image deblurring and image dehazing~\cite{restormer, daclip}. In this paper, we stick to a normal setting to focus on the effect of the prompt.

\begin{table*}[h]
    \centering
    \small
    \caption{Summary of datasets for various restoration tasks.}
    \newcolumntype{M}[1]{>{\centering\arraybackslash}p{#1}}
    \begin{tabular}{lllll}
    \shline
    Task & Training Set & Test Set & Details & Resampling ratio \\
    \hline
    Denoising & BSD400~\cite{BSD400}, WED~\cite{ma2016waterloo_wed} &  BSD68~\cite{martin2001database_bsd}, URBAN100~\cite{huang2015single_urban100} & \tabincell{l}{400 and 4744 pairs for training, \\ and 68 and 100 images for testing}   & $\times 3$ \\
    Deraining & Rain100L~\cite{yang2020learning_rain100} & Rain100L~\cite{yang2020learning_rain100} & 200 for training, 100 for testing & $\times 120$ \\
    Dehazing & SOTS~\cite{li2018benchmarking_SOTS} & SOTS~\cite{li2018benchmarking_SOTS} & 72135 for training, 500 for testing & $\times 1$ \\
    Deblurring & GoPro~\cite{gopro} & GoPro~\cite{gopro} & 2103 for training, 1111 for testing & $\times5$ \\
    Enhancement & LOL~\cite{lol} & LOL~\cite{lol} & 485 for training, 15 for testing & $\times 20$ \\
    \bottomrule[0.1em]
    \end{tabular}
    \label{tab:dataset_summary}
\end{table*}

\subsection{Network and Optimization Details}

\begin{figure}[htb]
    \centering    
    \includegraphics[width=1\linewidth]{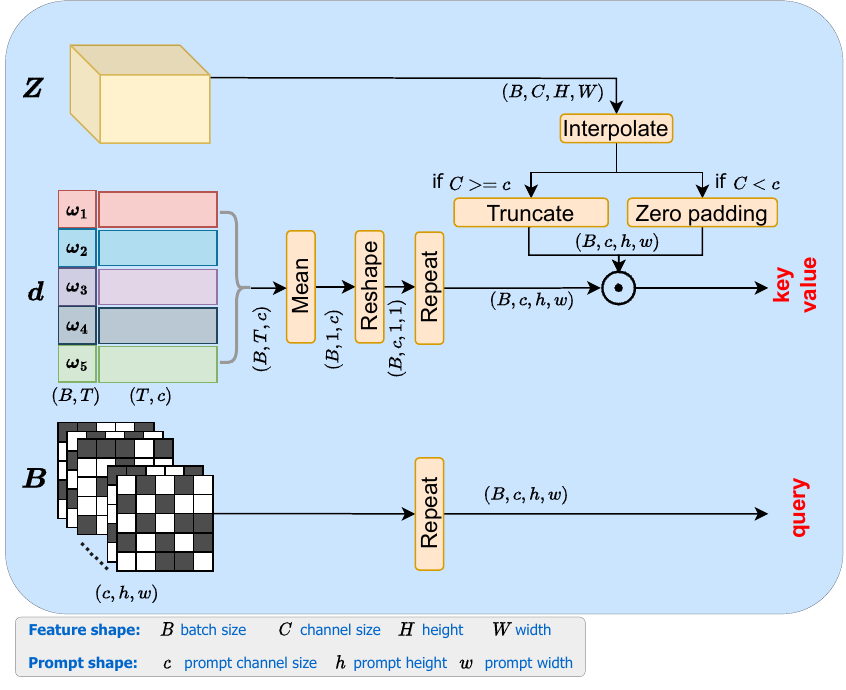}
    \caption{Detailed architecture for generating key, value, and query in our prompt-in-prompt learning.}
    \label{fig:pipdetails} 
\end{figure}

\paragraph{Details and configuration of \methodname.}
\cref{fig:pipdetails} illustrates the detailed architecture of \methodname before prompt-to-prompt interaction. To make sure \methodname works independently of the feature map shape, we align the feature map $\mat{Z}$ with the specific shape of prompts in this phase. Therefore, the prompt dimensions, $c$ for channels, $h$ for height, and $w$ for width, are treated as hyperparameters. The detailed configuration of these hyperparameters for PIP\textsubscript{Restormer} and PIP\textsubscript{NAFNet} are presented in Tabs.~\ref{tab:piprestormer} and \ref{tab:pipnafnet}, respectively.
For selective prompt-to-feature interaction, we adopt \cite{topkderain} to set an interval range for top-$m$ selection. This avoids either insufficient information or over-smoothed results. Specifically, we set $m$ at $\frac{C}{2}$, $\frac{2C}{3}$, $\frac{3C}{4}$, and $\frac{4C}{5}$ to create four distinct masks and corresponding outputs, respectively. These outputs are then individually scaled by a single learnable parameter. Finally, we sum these scaled outputs to get the final result of the selective cross-attention. 
\begin{table}[htb]
    \centering
    \small
    \caption{Configuration of hyperparameters for PIP\textsubscript{Restormer}. ``skip 1'' refers to the skip connection in the shadow layer and ``skip 3'' denotes the deepest ones. }
    \begin{tabular}{lcc}
    \shline
    layer & $c$ & $h\times w$  \\
    \hline
    skip 1        & 64   & $64\times64$  \\
    skip 2        & 128  & $32\times32$ \\
    skip 3        & 256  & $16\times16$  \\
    \shline
    \end{tabular}
    \label{tab:piprestormer}
\end{table}
\begin{table}[htb]
    \centering
    \small
    \caption{Configuration of hyperparameters for PIP\textsubscript{NAFNet}. ``skip 1'' refers to the skip connection in the shadow layer and ``skip 4'' denotes the deepest ones.}
    \begin{tabular}{lcc}
    \shline
    layer & $c$ & $h\times w$  \\
    \hline
    skip 1        & 64   & $64\times64$  \\
    skip 2        & 96  & $32\times32$ \\
    skip 3        & 128  & $16\times16$  \\
    skip 4        & 128  & $16\times16$  \\
    \shline
    \end{tabular}
    \label{tab:pipnafnet}
\end{table}

\paragraph{Detailed configuration of backbone models.}
We utilize Restormer~\cite{restormer} and NAFNet~\cite{nafnet} as the backbone models. 
Restormer is structured as a 4-level U-shape encoder-decoder network with 3 skip connections. We use the default settings of the official implementation, which is available at \url{https://github.com/swz30/Restormer}. This includes [4, 4, 6, 8] transformer blocks from shallow layer to deep layer, respectively. The basic dim of Restormer is 48 as default. \methodname is integrated into the three skip connections, with prompt dimensions set to [64, 128, 256] for skip connections, from the shallow layers to deep ones, as presented in \cref{tab:piprestormer}.
NAFNet is a 5-level U-shape encoder-decoder network with 4 skip connections. We utilize the official code from \url{https://github.com/megvii-research/NAFNet}, adopting the default architecture designed for image deblurring, which is more balanced in the encoder and decoder. This includes [2, 2, 4, 8] blocks for the encoder, 12 blocks for the middle bottleneck, and [2, 2, 2, 2] blocks for the decoder, respectively. The width of NAFNet is set to 64 as default. Our \methodname contains four skip connections, the prompt dim is set to [64, 96, 128, 128] to reduce the computational and memory costs, as shown in \cref{tab:pipnafnet}.

\paragraph{Training details.}
\methodname is trained alongside the backbone models from scratch for 200 epochs. We employ the cosine annealing schedule with a linear warm-up~\cite{sgdr} to control the learning rate. Initially, the learning rate linearly increases to $5 \times 10^{-4}$ over the first 10 epochs. The learning rate then gradually decreases according to a cosine annealing manner, reaching 0 at the 150-th epoch, and resumes its increase in a cosine manner until the end of the training.

\section{Additional Results}
\label{sec:results}


\subsection{Control by Degradation-aware Models}

\methodname is designed only to focus on enhancing the restoration performance rather than automatically recognizing different tasks. This is due to the complicated nature of degradation and various requirements in application. For instance, a user may want to remove noise from a low-light image without increasing brightness.  
However, we argue that controlling \methodname with degradation-aware models is applicable and can be easily implemented. 
For simplicity, we just employ the most commonly used classification backbones, including VGG-16, ResNet-34, and ResNet-34 built with Fourier convolution~\cite{FFC}. 
We use Fourier convolution based on the observation that some degradations, such as noise and rain, primarily affect high-frequency details, which can be excessively smoothed by spatial convolution. Specifically, we use a ResNet-34 built with fast Fourier convolution from the official implementation at \url{https://github.com/pkumivision/FFC}. 

All three classification backbones are trained on a task to classify the five different types of degradation. The training is conducted on 4 RTX4090 GPUs, with a batch size of 128 and an initial learning rate of 0.01. The Adam optimizer is employed with $\beta_1=0.9$ and $\beta_2=0.999$, respectively. The learning rate is reduced by half every 30 epochs. We maintain consistent data preparation and train the model for a total of 200 epochs. 

\begin{table}[h]
    \centering
    \small
    \caption{Accuracy of degradation-aware models in degradation classification.}
    \begin{tabular}{lrrr}
    \shline
    Task & VGG16 & ResNet-34 & ResNet-34 w/ Fourier  \\
    \hline
    Noise       & 0.00   & 98.04 & 100.00 \\
    Rain        & 0.00   & 99.00 & 100.00 \\
    Haze        & 100.00 & 95.60 &100.00 \\
    Blur        & 0.00   & 100.00   &100.00 \\
    Low light   & 0.00   & 100.00   & 100.00 \\
    \shline
    \end{tabular}
    \label{tab:control}
\end{table}

Tab.~\ref{tab:control} presents the results. 
Unsurprisingly, we find that a simple degradation-aware model built with fast Fourier convolution can easily obtain $100\%$ accuracy on all five tasks, demonstrating that it can \emph{easily control \methodname without a performance drop}. We also observe that ResNet-34 also achieves a good performance while VGG-16 collapses in the training. The reason is that VGG contains a series of downsampling layers without enough skip connections to preserve the low-level details. In comparison, ResNet can better fit the task of degradation classification.

\subsection{Training with Multiple Prompts}

In real-world scenarios, a key challenge is that images can suffer from multiple types of degradation simultaneously. Existing datasets seldom contain such cases. 
To investigate the capability of tackling multiple degradations, one straightforward strategy is to simulate various degradations on a clean image~\cite{gopro, tpami_degradation}. However, accurately replicating real-world conditions remains challenging, especially for degradation like haze and blur. This also presents a significant challenge for restoration models, as they must adapt to a wide range of degradation combinations and permutations~\cite{mix_restore, mix_cvpr}.
\begin{figure}[t]
    \centering    
    \includegraphics[width=0.8\linewidth]{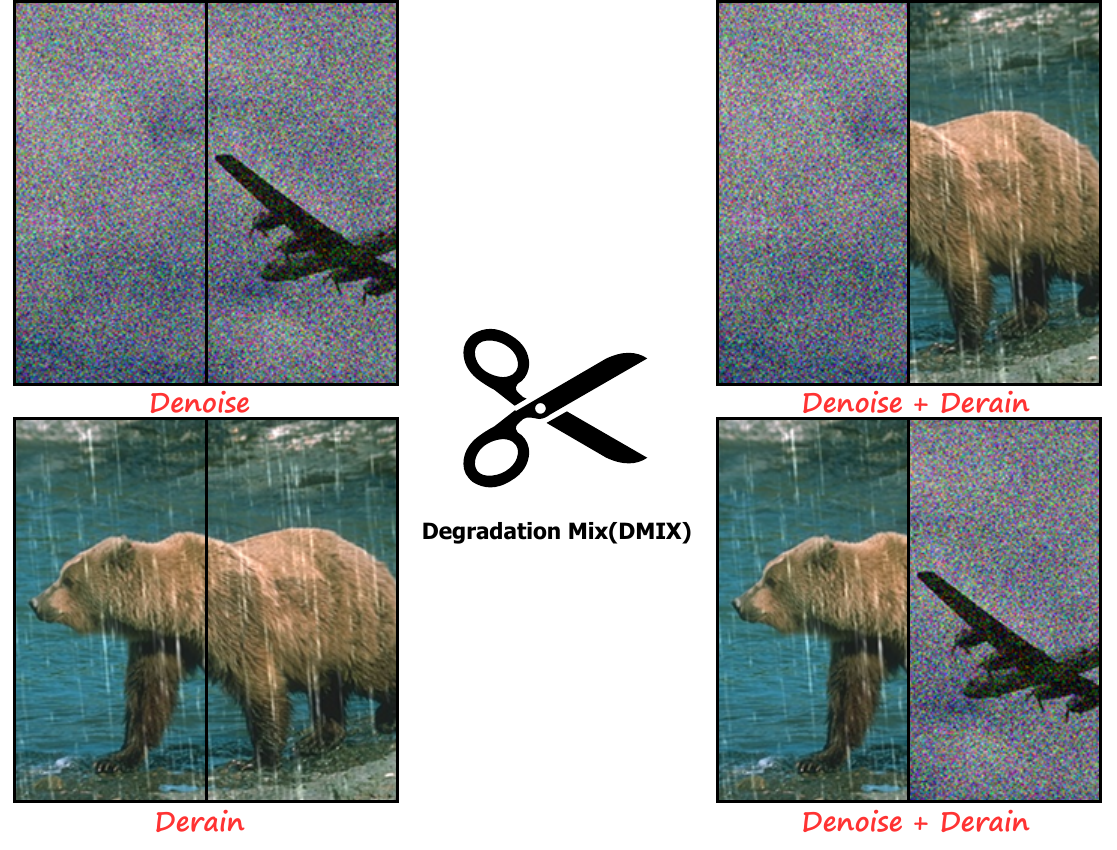}
    \caption{Illustration of Degradation-Mix data augmentation in the horizontal axis. }
    \label{fig:dmix} 
\end{figure}

\begin{figure}[t]
    \centering    
    \includegraphics[width=1\linewidth]{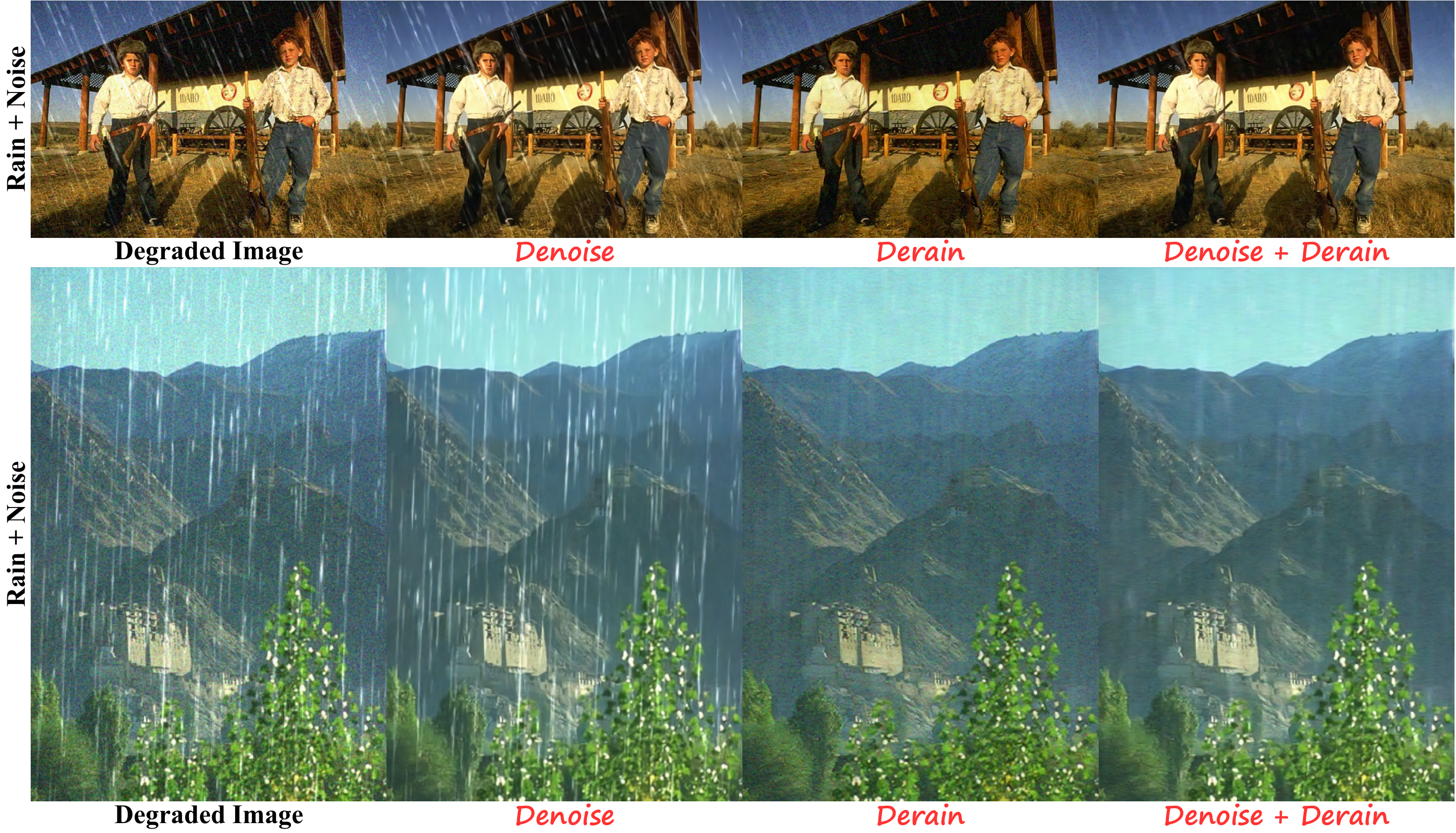}
    \vspace{-10pt}
    \caption{Visualization of \methodname restoration with different prompts. }
    \label{fig:dmix_result} 
\end{figure}

Despite the challenges of the task, we explore the potential of training the restoration model using multiple prompts with data augmentation.  
Inspired by augmentation techniques that blend multiple images into one~\cite{cutmix}, or apply various enhancements to a single image~\cite{cutonce}, and can improve the classification performance, we design an augmentation strategy for training called Degradation-Mix (DMIX), as shown in Fig.~\ref{fig:dmix}. 
Similar to Cut-Mix~\cite{cutonce}, we first split all images in a mini-batch into two halves, either horizontally or vertically. These halves are then randomly shuffled to create new images with combined degradations. Concurrently, we mix the corresponding degradation labels to align with the degradation-aware prompt.
By training \methodname with DMIX, we observe that \methodname exhibits improved generalization in handling images with mixed degradations, as seen in Tab.~\ref{tab:suppl_ood}. We also find that it produces better visual results by using multiple prompts at the same time, as shown in Fig.~\ref{fig:dmix_result}. However, we also observe some remaining artifacts and the quality of the restored image was not optimal. 
These results underline the flexibility and potential of the proposed prompt-in-prompt learning. 
Future research could focus on utilizing multiple prompts collaboratively addressing mixed degradations.

\begin{table}[t]
    \centering
    \small
    \caption{Result of \methodname trained with multiple prompts. Metrics are presented in [PSNR (dB) / SSIM].}
    \begin{tabular}{lccc}
    \shline
    Method & Rain100L + Noise & Haze + Noise   \\
    \hline
    PIP\textsubscript{Restormer}         & 26.50/0.771 & 16.58/0.754  \\
    PIP\textsubscript{Restormer} w/ DMIX & 26.79/0.815 & 20.61/0.741  \\
    \shline
    \end{tabular}
    \label{tab:suppl_ood}
\end{table}

\begin{table}[t]
    \centering
    \small
    \caption{Generalization to real-world denoising and medical image denoising. Metrics are presented in [PSNR (dB) / SSIM].}
    \begin{tabular}{lccc}
    \shline
    Dataset & AirNet & PromptIR & PIP\textsubscript{Restormer}  \\
    \hline
    SIDD~\cite{SIDD_denoise} & 23.66/0.356 & 24.32/0.378 & 24.12/0.362 \\
    AAPM~\cite{aapm}  & 34.59/0.936 & 34.61/0.9432 & 34.67/0.919 \\
    \shline
    \end{tabular}
    \label{tab:gen}
\end{table}

\subsection{Generalization on Real-World Data}
The results in Tab.~\ref{tab:gen} reveal that all three methods struggle with zero-shot performance on unseen types of degradation, such as noise in the SIDD~\cite{SIDD_denoise} dataset that consists of real camera noise rather than Gaussian noise. 
When generalizing to medical images in ``2016 NIH-AAPM-Mayo Clinic Low-Dose CT Grand Challenge'' (AAPM) dataset~\cite{aapm} affected by Gaussian noise with $\sigma = 25$, all three methods perform relatively better. 

Although \methodname performs relatively better than compared methods in these scenarios, we found that the generalizability is still \emph{mainly determined by the backbone model}. 
Conventional restoration backbones are less effective with new degradation types, despite being efficient in inference. On the other hand, popular diffusion-based methods show better adaptability across various tasks~\cite{stablediffusion, stablediffusion_inpainting} while requiring more time for inference and higher computational costs. Choosing an appropriate backbone model involves a trade-off.
Hence, the challenge of adapting a trained model to new types of degradation remains an open question and could be a promising direction for future research.

\subsection{More Visual Results}

\begin{figure*}[!htb]
    \centering    
    \includegraphics[width=1\linewidth]{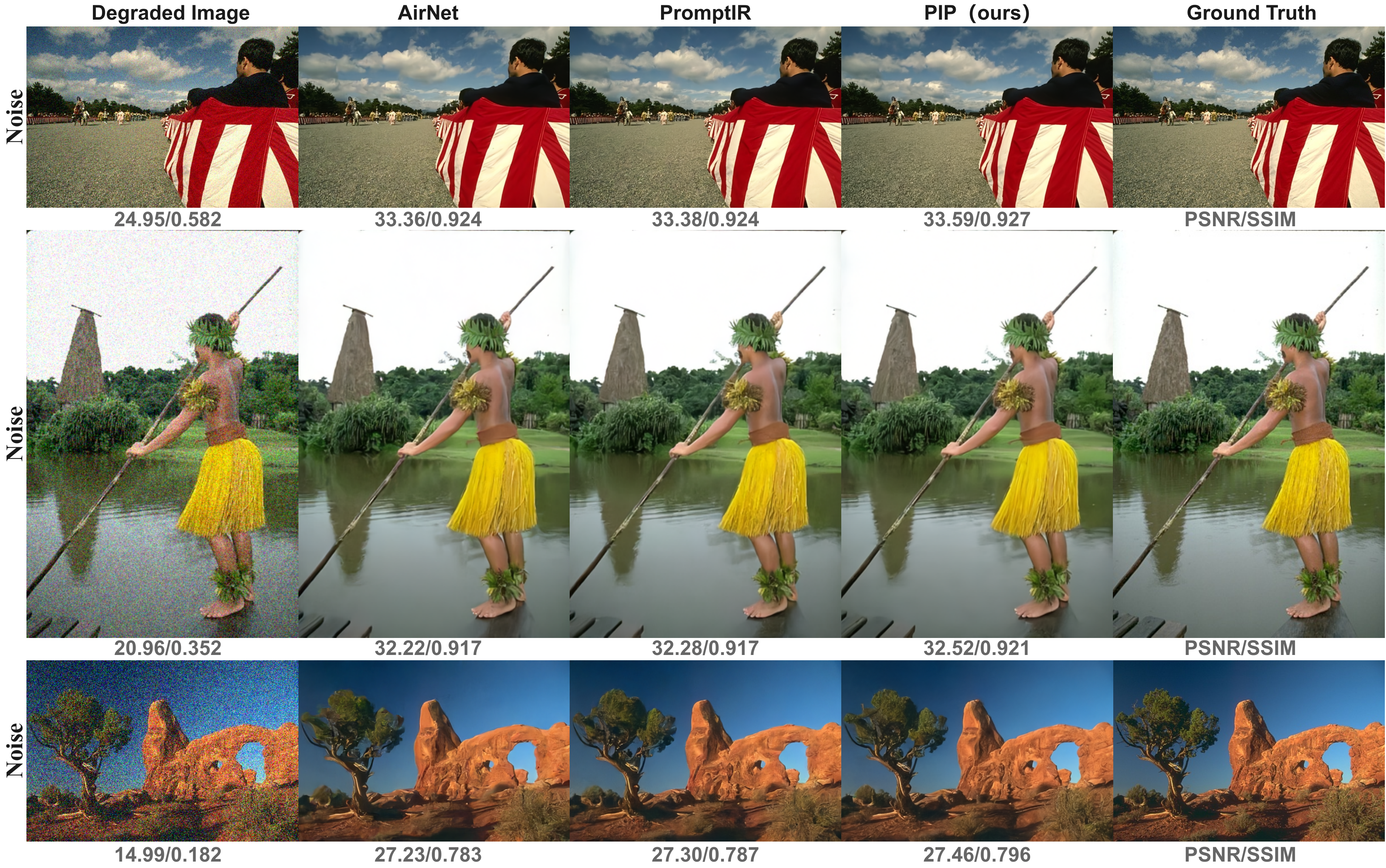}
    \caption{Visual comparison of universal methods on image denoising with a Gaussian noise level $\sigma = 15, 25, 50$, from the first to the third row, respectively. }
    \label{fig:suppl_denoise} 
\end{figure*}

\begin{figure*}[!htb]
    \centering    
    \includegraphics[width=1\linewidth]{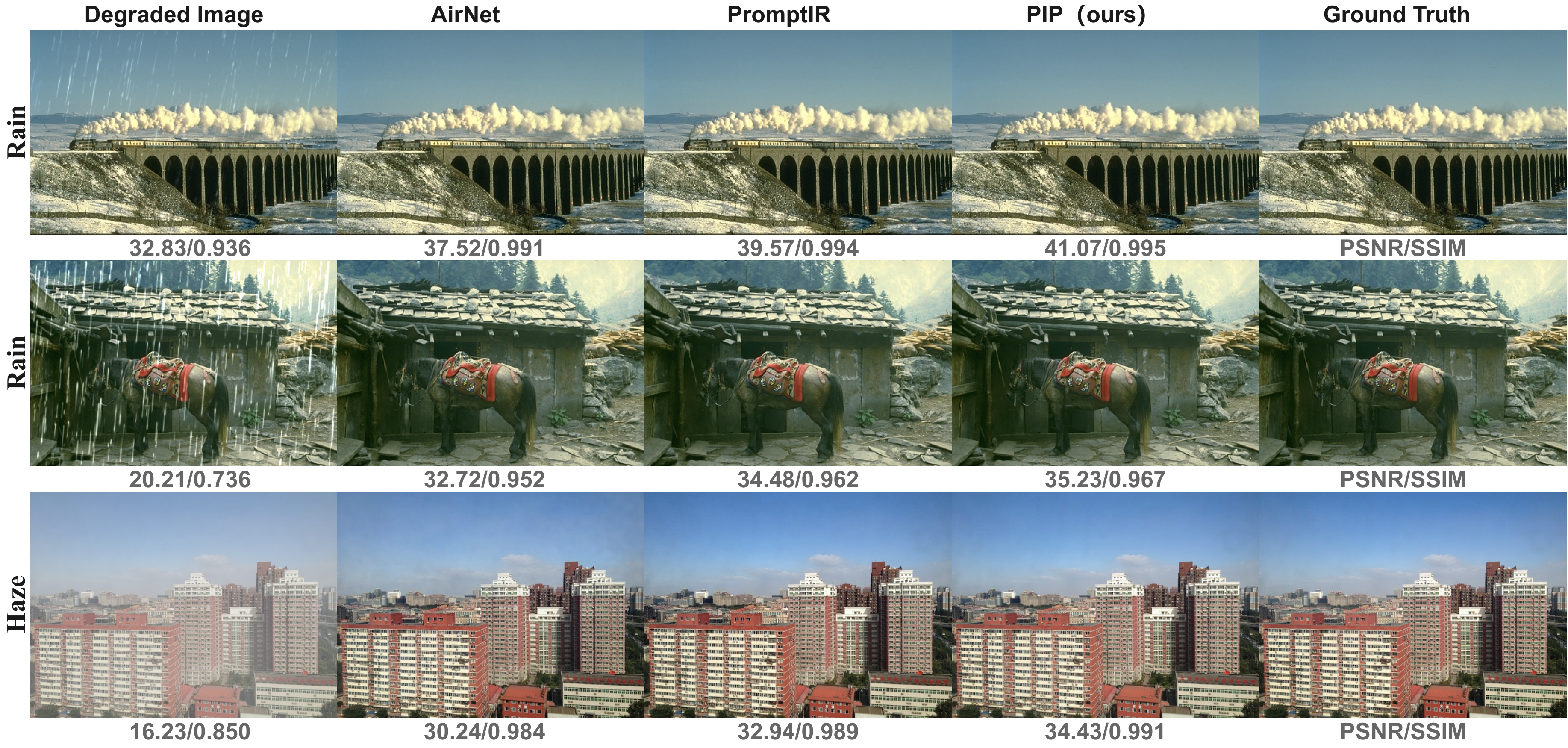}
    \caption{Visual comparison of universal methods on image deraining and dehazing. }
    \label{fig:suppl_derainhaze} 
\end{figure*}

\begin{figure*}[!htb]
    \centering    
    \includegraphics[width=1\linewidth]{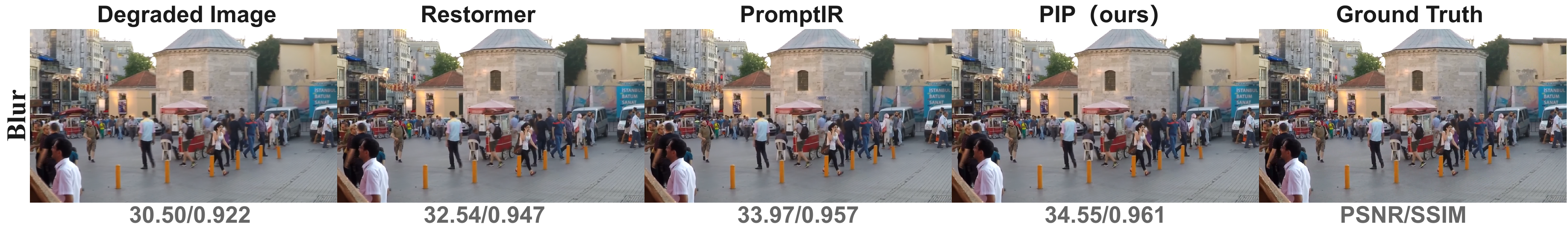}
    \includegraphics[width=1\linewidth]{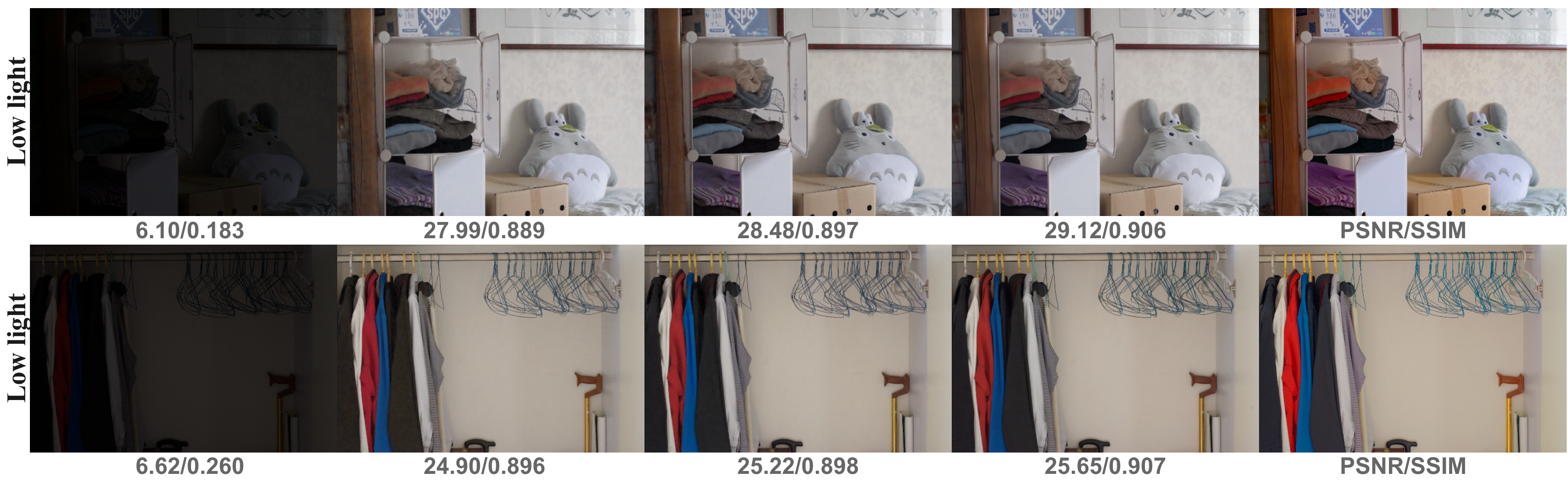}
    \caption{Visual comparison of universal methods on image deblurring and low light enhancement. }
    \label{fig:suppl_deblurlowlight} 
\end{figure*}

Figs.~\ref{fig:suppl_denoise},~\ref{fig:suppl_derainhaze}, and~\ref{fig:suppl_deblurlowlight} show more visualization results of \methodname and other universal restoration methods. Generally, we find that \methodname achieves optimal visual effect.

\section{More Discussion}
\label{sec:discuss}
In summary, the proposed prompt-in-prompt learning effectively enhances existing backbones in universal image restoration and surpasses recent methods with only a slight increase of computational cost. Prompt-in-prompt learning merges the advantages of high-level degradation-aware prompt and low-level basic restoration prompt, thereby offering clear interpretability and flexible control options, whether through human intervention or degradation-aware models. We emphasize that both \methodname and designed modules can be easily adapted to other backbone models and tasks in low-level vision.

\paragraph{Limitations.} First, \methodname still introduces some computational costs in both training and inference phases, although these are substantially lower than those of comparative methods. Second, we observe that \methodname does not significantly enhance the generalization capabilities of models to unknown degradations. 
Although this may be attributed to the backbone models and limited parameters, it is worth investigating whether prompt-in-prompt learning can be beneficial for large-scale restoration models. Furthermore, exploring how to use prompts to boost the zero-shot performance of restoration models is also a worthwhile field.

\end{document}